\documentclass{article}

\usepackage{microtype}
\usepackage{graphicx}
\usepackage{subfigure}
\usepackage{booktabs} % for professional tables
\usepackage{amsfonts}
\usepackage{amsmath}
\usepackage{amssymb}
\usepackage{algpseudocode}
\usepackage[table]{xcolor}
\usepackage{balance}
\usepackage{hyperref}

\newcommand{\name}{LoRazor}

\usepackage[accepted]{mlsys2025}

\mlsystitlerunning{LoRazor: Trimming Adapter Execution for Efficient Multi-LoRA Serving}

\begin{document}

\twocolumn[
\mlsystitle{LoRazor: Trimming Adapter Execution for Efficient Multi-LoRA Serving}

% It is OKAY to include author information, even for blind
% submissions: the style file will automatically remove it for you
% unless you've provided the [accepted] option to the mlsys2025
% package.

% List of affiliations: The first argument should be a (short)
% identifier you will use later to specify author affiliations
% Academic affiliations should list Department, University, City, Region, Country
% Industry affiliations should list Company, City, Region, Country

% You can specify symbols, otherwise they are numbered in order.
% Ideally, you should not use this facility. Affiliations will be numbered
% in order of appearance and this is the preferred way.
\mlsyssetsymbol{equal}{*}

\begin{mlsysauthorlist}
\mlsysauthor{Xiang Li}{}
\mlsysauthor{Pengcheng Wang}{}
\mlsysauthor{Azam Ikram}{}
\mlsysauthor{Saurabh Bagchi}{}
% \mlsysauthor{Fiuea Rrrr}{to}
% \mlsysauthor{Tateu H.~Yasehe}{ed,to,goo}
% \mlsysauthor{Aaoeu Iasoh}{goo}
% \mlsysauthor{Buiui Eueu}{ed}
% \mlsysauthor{Aeuia Zzzz}{ed}
% \mlsysauthor{Bieea C.~Yyyy}{to,goo}
% \mlsysauthor{Teoau Xxxx}{ed}
% \mlsysauthor{Eee Pppp}{ed}
\end{mlsysauthorlist}
\begin{center}
    Purdue University
\end{center}

% \mlsysaffiliation{pur}{Purdue University}
% \mlsysaffiliation{goo}{Googol ShallowMind, New London, Michigan, USA}
% \mlsysaffiliation{ed}{School of Computation, University of Edenborrow, Edenborrow, United Kingdom}

% \mlsyscorrespondingauthor{Cieua Vvvvv}{c.vvvvv@googol.com}
% \mlsyscorrespondingauthor{Eee Pppp}{ep@eden.co.uk}

% You may provide any keywords that you
% find helpful for describing your paper; these are used to populate
% the "keywords" metadata in the PDF but will not be shown in the document
\mlsyskeywords{Machine Learning, MLSys}

\vskip 0.3in

\begin{abstract}
LoRA offers a parameter-efficient adaptation strategy, supporting multi-user workloads via customized adapters on a shared backbone.
However, serving concurrent requests with different adapters introduces substantial computation and memory-management overhead. We ask the research question: Do all requests need their adapters throughout the whole generation process? 
Our motivational study reveals that employing LoRA adapters at every decoding step is unnecessary; releasing them early has limited impact on accuracy while substantially accelerating generation. Although skipping adapters early offers theoretical workload reduction, translating these computational savings into concrete throughput gains requires tailored GPU kernels.
We present LoRazor, a multi-LoRA serving framework that combines early adapter release with active-row-aware GPU kernels. LoRazor determines when to permanently release a request’s adapter and continue decoding with the backbone alone. 
By dynamically resizing LoRA kernel launches as adapters are released, LoRazor trims idle GPU blocks.
Across two backbones and four tasks, LoRazor achieves up to 3.02$\times$ request throughput and 2.53$\times$ output-token throughput over vLLM multi-LoRA, with a mean relative task-accuracy decrease of 2.89\%.
\end{abstract}
]

% this must go after the closing bracket ] following \twocolumn[ ...

% This command actually creates the footnote in the first column
% listing the affiliations and the copyright notice.
% The command takes one argument, which is text to display at the start of the footnote.
% The \mlsysEqualContribution command is standard text for equal contribution.
% Remove it (just {}) if you do not need this facility.

%\printAffiliationsAndNotice{}  % leave blank if no need to mention equal contribution
%\printAffiliationsAndNotice{\mlsysEqualContribution} % otherwise use the standard text.

\section{Introduction}
\label{sec:introduction}
% LLM good at doing normal task, domain specifica need specializaiton....
% provider : huggingface amazon bedrock
Large language models (LLMs) support a broad range of tasks, from answering questions to generating code. However, their general-purpose capabilities may fall short when applications require specialized domain knowledge, task-specific reasoning, or precise output formats. Low-rank adaptation (LoRA)~\cite{hu2022lora} provides a parameter-efficient way to specialize these models by freezing the pretrained backbone and learning low-rank, task-specific weight updates~\cite{kong2024customizing,zhang2024personalization,li2026embodied}. A provider (e.g., HuggingFace) can then serve many such adapters on a shared backbone rather than maintain a full model copy for each task. Despite the compact size of individual adapters making it possible to fit more adapters in GPU memory, serving those adapters concurrently still leads to resource contention, which degrades serving performance.

Prior work addresses this through efficient multi-adapter GPU execution and memory management~\cite{chen2024punica,slora,dlora}, shared adapter representations~\cite{kopiczko2024vera,cts,li2026pin}, and prompt-only adaptation~\cite{ma2025prad,lanpouthakoun2026preft}. 
We examine a complementary question: how long does each request need its assigned adapter during the decoding stage? Specifically, can the request stop using its adapter before generation finishes and use the backbone alone to complete the response? We refer to this permanent adapter deactivation for a specific request as \emph{adapter release} and investigate its quality-efficiency trade-off.

% Azam: I think the "release" refer to the method where we do certain number of decodes with the adapter and the rest are done only on the backbone model. But simply using the term release dooesn't imply that following decodes on the backbone. Might be better to establish what we mean by release before making a claim here.
We conduct a comprehensive motivational study on four datasets and two backbones.
Our experiments (Sec~\ref{sec:motivation}) show that the impact of adapter release on task accuracy varies across tasks and backbones. 
% In some settings, applying an adapter during prefill and several initial decoding iterations preserves much of the adapted model's accuracy, even when subsequent decoding uses only the backbone. 
Under certain circumstances, applying the adapter during prefill and several initial decoding iterations already preserves much of the adapted model's accuracy.
In others, early release may substantially degrade quality. These observations motivate our selective adapter release but also raise two questions:
\emph{When should each request stop using its adapter, and how can the serving system translate these decisions into higher serving throughput?}

Deciding when to release an adapter requires assessing its effect on the remaining generation. Because adapter release changes both subsequent token generation and the KV cache, its impact on final task quality may not be apparent immediately.
% Azam: How is a release policy connected to decoding overhead? What decoding overhead are we referring to here? 
Thus, a practical release policy must account for the downstream effects while adding little overhead to the batch execution time. Furthermore, adapter release alone does not ensure higher serving efficiency.
% Azam: I think this is the decoding overhead the above sentence talks about. If true, probably structure it better by introducing it first and then tieing it back to the overhead that would allow us get execution savings.
As requests release their adapters, fewer tokens need to compute LoRA updates in each batch, but the existing LoRA computation kernel~\cite{kwon2023efficient} remains sized for the full padded batch, leaving excess GPU blocks that incur block-scheduling and metadata-access overhead without performing useful matrix computation. Realizing the benefits of adapter release therefore requires adapting kernel launches to the adapter workload at every iteration.

To solve the aforementioned problems, \textit{we present \name, a multi-LoRA serving framework that combines selective adapter release with
workload-aware kernel execution.} \name\ consists of two components: \textsc{StepTrim} and \textsc{GridTrim}.
% Azam: Mention section number or skip it from RowPulse too
\textsc{StepTrim} (Sec~\ref{sec:steppulse}) uses a release policy to decide when a request releases its adapter and continues decoding with the backbone alone.
% Azam: following is a good defination of the term release, introduce it earlier.
% d. We train the predictor using paired continuations (Sec~\ref{sec:steppulse}) to estimate the risk of quality degradation. When the predicted risk falls below a selected threshold, the request switches to backbone-only decoding while retaining its existing KV cache.
% Azam: I wonder if we can also claim that by doing so we allow the backbone to execute larger decoding batches which are known to produce higher throughput. If we do say that, make sure we have an experiment to show it too.
This reduces the number of tokens processed by each adapter, creating opportunities to shrink the corresponding GPU launch grids.
\textsc{RowTrim} (Sec~\ref{sec:rowpulse}) exploits these opportunities by adapting grouped LoRA kernel launches to the remaining adapter workload.
% Azam: The following sentence is difficult to understand. It says bounds twice which feels structurally awkward. I also don't know what bounds refer to here
%replacing full-batch row bounds with bounds determined by active adapter group sizes. %It supports CUDA-graph replay through a small set of captured launch variants.
% Azam: What are active group sizes?
Together, they reduce adapter computation and translate reductions in active LoRA group sizes into kernel execution savings. 

We implement \name\ in vLLM~\cite{kwon2023efficient} and evaluate it on two widely used backbone LLMs, Llama-3.2-3B and Mistral-7B, with adapters spanning four tasks from four fields including mathematics, coding, medical and science. \name\ achieves throughput speedups of $2.56\times$ and $3.02\times$, respectively, over vLLM multi-LoRA serving (Punica derived LoRA kernels), with an average accuracy decrease of 1.82 percentage points, which is only 2.89\% relative to the full-LoRA execution. Our contributions are:
\begin{itemize}
    % Azam: The above introduction doesn't say anything about any characterization. Might be better to reword the paragraph where we observed the problem to be the characterization.
    \item We identify and systematically exploit opportunities to reduce multi-LoRA decoding cost through selective adapter release and reduction of inactive GPU blocks.
    % We identify and systematically exploit opportunities
    % Azam: What are active tokens? Don't introduce a new term here
    \item We design and develop \name, which consists of \textsc{StepTrim} for early selective adapter release and \textsc{RowTrim} for adapting grouped LoRA kernel launches to active adapters.
    \item We integrate \name\ into the state-of-the-art vLLM serving engine and evaluate its quality-throughput trade-offs using diverse tasks and backbones to demonstrate its efficiency.
\end{itemize}
\section{Motivation}
\label{sec:motivation}

\subsection{Adapter Diversity Increases Decode Latency}
\label{sec:motivation-diversity}

\begin{figure}[h]
  \centering
  \includegraphics[width=\columnwidth]{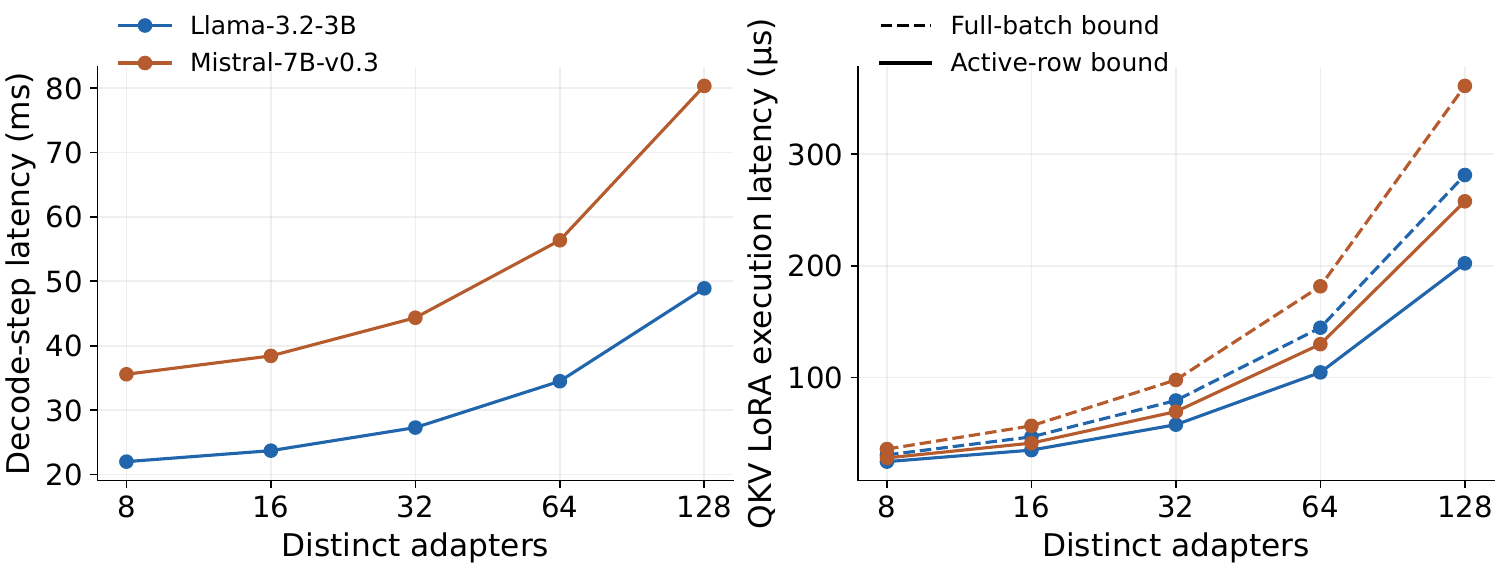}
  \vspace{-4 mm}
  \caption{(a) For a fixed batch of 256 requests, per-step latency increases with the number of distinct adapters. (b) Using active row bound reduces the LoRA execution latency. }
  \label{fig:serving-motivation}
\end{figure}

A multi-LoRA batch contains requests that need to be assigned to different adapters. Existing framework~\cite{kwon2023efficient} provision GPU work using the full batch-wide row dimension for each active adapter group, even when an adapter is assigned only a small subset of requests. Consequently, higher adapter diversity creates additional GPU work that is not proportional to the number of requests that are tied to each adapter. Figure~\ref{fig:serving-motivation}(a) demonstrates this effect for a batch of 256 requests. Increasing the number of distinct active adapters from 8 to 128 raises the median per-step decode latency from 21.91 ms to 48.82 ms for Llama-3.2-3B and from 35.72 ms to 80.04 ms for Mistral-7B-v0.3. 

Figure~\ref{fig:serving-motivation}(b) shows the cost of oversized kernel launches. The full-batch bound provisions 256 rows for every adapter regardless of how many rows are actually being used. We instead use an \emph{active-row bound}: a shared row bound determined by the largest number of tokens assigned to any active adapter, rounded up to a supported launch size. 
% SB (10/29/26): The above sentence is packing too much into one. What is an active adapter? What is launch size (GPU architecture allows these many adapters to be in the memory concurrently)? 
This reduces the row tiles provisioned for each adapter while retaining all required adapter computation. 
% SB (10/29/26): The above seems impossible - to reduce the number of rows and yet retain all computation.
We compare LoRA execution latency when using these two launch bounds. At 128 adapters, the active-row bound reduces latency by 28.1\% on Llama-3.2-3B and 28.6\% on Mistral-7B-v0.3. These results demonstrate the cost of provisioning row tiles beyond the active adapter workload and motivate \emph{GridTrim}, which adapts kernel launch bounds to active adapter-group sizes.

\subsection{LoRA Is Not Required at Every Decode Step}
\label{sec:motivation-skipping}

\begin{figure}[ht]
  \centering
  \includegraphics[width=\columnwidth]{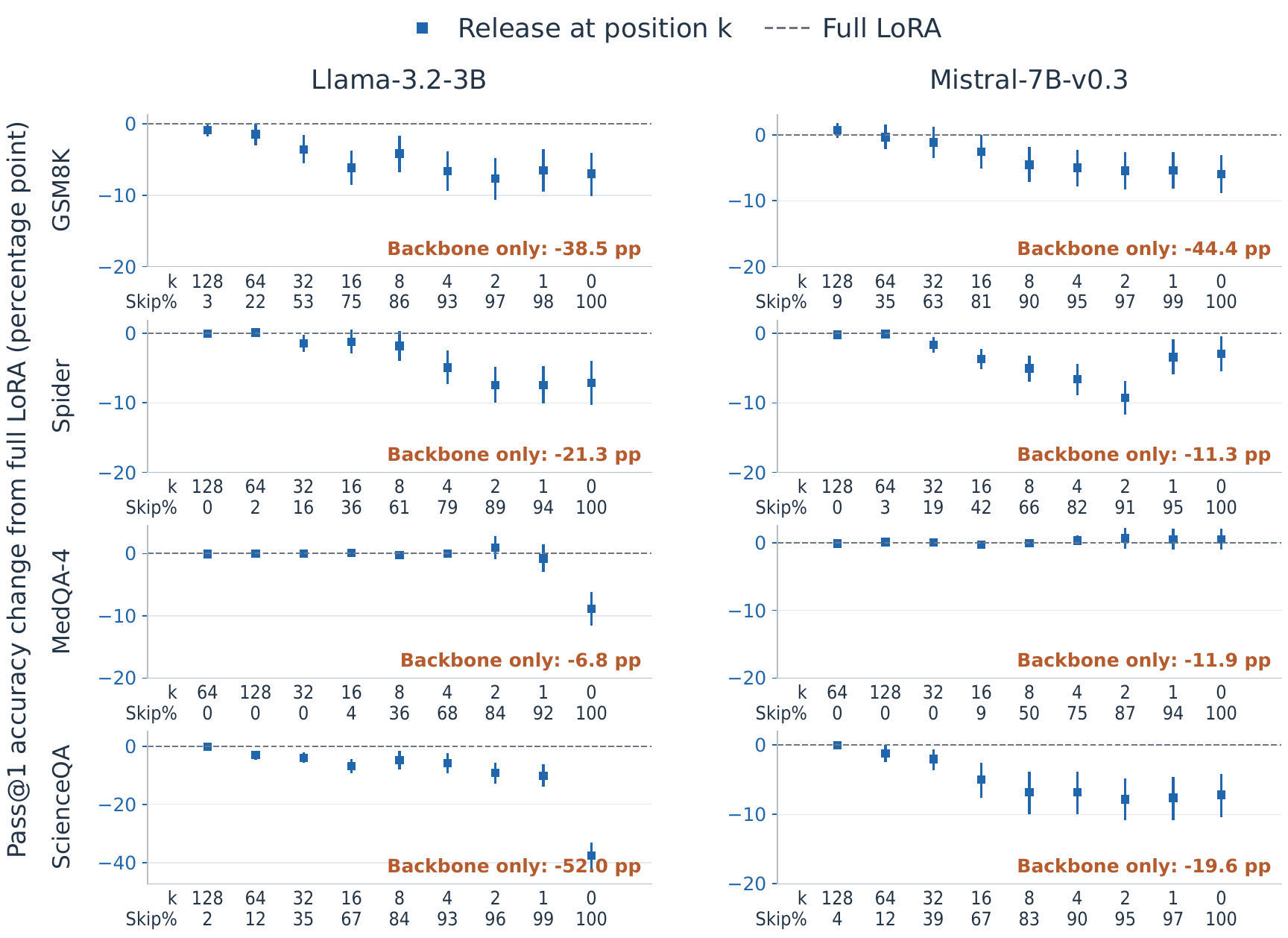}
  \vspace{-4 mm}
  \caption{The quality impact of decode-time LoRA adapter release varies across tasks and backbones. The best release position is task- and backbone-dependent.}
  \label{fig:decode-skipping}
  \vspace{-2mm}
% SB (10/29/26): The above figure would not be understandable by any reviewer. My suggestion is that we do this in a table. There will be 4 (tasks) X 2 (models) subtables. For each subtable, one row for each value of k. The columns are time reduction and quality reduction. The full LoRA will be one upper bound number for each subtable. Also, the caption should explain what k is. 
\end{figure}

Existing system applies a request's adapter during prefill and every subsequent decode step. However, the adapter's contribution is not equally important throughout the generation. 
We study this behavior by disabling LoRA during decoding steps while retaining it throughout prefill.
% We study this behavior by disabling LoRA during part of 
% % SB (10/29/26): Crucial addition "part of"
% decoding while retaining LoRA throughout prefill.

Figure~\ref{fig:decode-skipping} evaluates adapter releases at fixed positions. The adapter remains enabled during prefill and the first \(k\) decoding steps, after which it is disabled for the remainder of generation. Skip (\%) denotes the fraction of decoding steps that omit adapter, relative to applying an adapter throughout decoding. We report test-accuracy changes relative to using an adapter throughout generation and the mean of fraction of decode adapter calls skipped across requests.
The quality-efficiency trade-off varies substantially across tasks and backbones. On MedQA-4, releasing after one decode step skips 92.0\% of decode adapter calls on average for Llama-3.2-3B while changing accuracy by $-0.63$ percentage points. In contrast, early release causes larger accuracy losses on GSM8K and ScienceQA tasks. Therefore, a fixed release point does not preserve quality uniformly. Furthermore, retaining LoRA for more decoding steps also does not always improve accuracy. Mistral-Spider exhibits a nonmonotonic trend: accuracy decreases from $k=0$ to $k=2$, then generally recovers as release is delayed. A close examination suggests that the first few adapter-guided steps can produce SQL prefixes that the backbone struggles to complete correctly after release. Releasing immediately after prefill instead lets the backbone generate both the initial SQL tokens and their continuation.
% SB (10/29/26): The above reasoning is critical as it explains a non-intuitive observation (intuition is larger is k, higher should be accuracy). But I do not understand the explanation --- "toward a SQL prefix ... continuation". 
Delaying release further generally improves accuracy. Our study in Appendix~\ref{app:prefill} shows disabling LoRA during prefill causes substantially larger accuracy losses, so we retain adapters throughout prefill and focus on release during decoding.

Together, we identify two opportunities to improve multi-LoRA serving efficiency: avoiding unnecessary GPU blocks for active adapters and shortening adapter execution when task accuracy permits. These findings motivate \name, where \textsc{StepTrim} decides when to permanently release an adapter and \textsc{GridTrim} resizes grouped LoRA kernel launches according to the remaining active adapter rows.
\section{Method}

\begin{figure*}[h]
  \includegraphics[width=\textwidth]{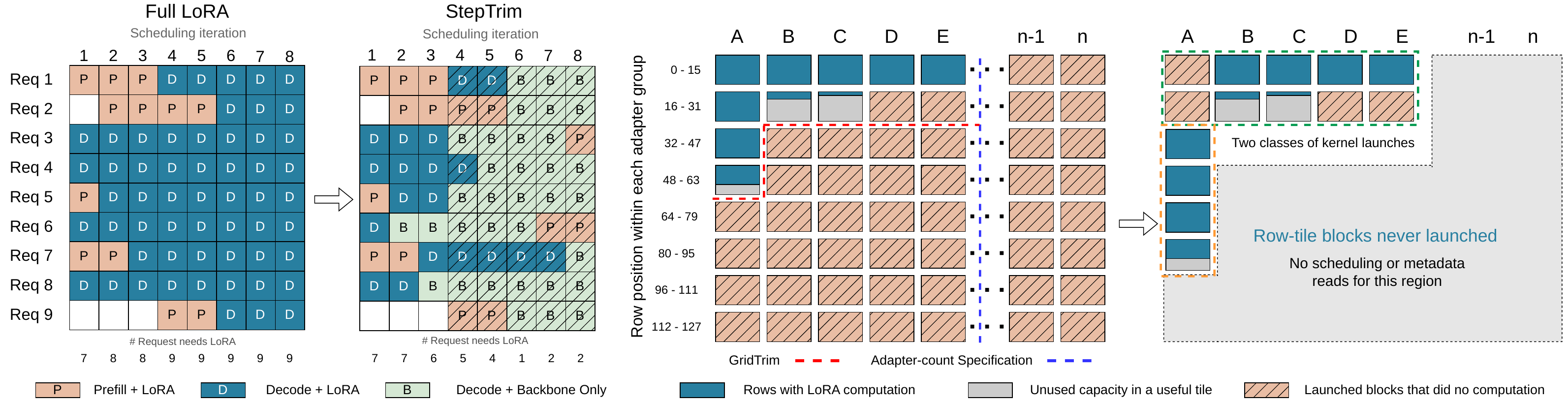}
  \vspace{-4mm}
  \caption{Overview of \name. Left: StepTrim retains LoRA during prefill and releases adapters during decoding. Requests then continue with the backbone alone and may terminate at different steps from full-LoRA execution. Right: GridTrim reduces the row dimension of grouped LoRA launch grids, complementing vLLM’s adapter-count specialization. Two-class expand execution handles imbalanced adapter-group sizes.}
  \label{fig:lorapulse}
\end{figure*}

\subsection{GridTrim: Active-Row LoRA Execution}
\label{sec:rowpulse}

\textsc{GridTrim} matches LoRA kernel launches to active adapter group sizes rather than the full batch. It uses a shared active-row bound for shrink and optionally separates expand into small- and large-class launches.

\subsubsection{Identifying Redundant GPU Blocks}
\label{sec:rowpulse-overlaunch}

Let $X\in\mathbb{R}^{M\times H}$ contain a linear layer's input activations, where $M$ is the padded batch row count and $H$ is the input hidden dimension. Adapter $a$ serves row indices $\mathcal{I}_a$, with $m_a=|\mathcal{I}_a|$, and $\mathcal{A}=\{a:m_a>0\}$ denotes the active adapters. For LoRA weights $A_a\in\mathbb{R}^{H\times r_a}$ and $B_a\in\mathbb{R}^{r_a\times D}$, the adapter computes
\begin{align}
    U_a &= X[\mathcal{I}_a]A_a,
    &&\text{(shrink)}
    \label{eq:rowpulse-shrink}\\
    \Delta Y_a &= U_aB_a,
    &&\text{(expand)}
    \label{eq:rowpulse-expand}
\end{align}
where $r_a$ is the adapter rank and $D$ is the output width. The correction $s_a\Delta Y_a$ is added to the backbone output at rows $\mathcal{I}_a$, with adapter scaling factor $s_a$.

For row-tile height $B_M$, the baseline~\cite{kwon2023efficient} launches $\lceil M/B_M\rceil$ row tiles per adapter slot, although only $\lceil m_a/B_M\rceil$ tiles perform matrix computation. Remaining tiles read the group's row count and exit, still incurring scheduling and metadata-access overhead (Figure~\ref{fig:lorapulse} right). For example, with $M=256$ and $B_M=16$, an adapter serving 10 rows needs one row tile but receives sixteen.

Let $L$ be the number of adapter slots provisioned by the launch. The baseline provisions $L\lceil M/B_M \rceil$ row tiles. GrdiTrim aims to reduce this by provisioning GPU kernel launches according to the active adapter request assignment in each batch.

\subsubsection{Specializing the Row Bound}
\label{sec:rowpulse-bound}

The full batch size is a conservative bound on every adapter's workload. A tighter bound is the largest active adapter group, $m_{\max}=\max_{a\in\mathcal{A}}m_a$, in the current batch. \textsc{GridTrim} selects a common row bound $R$ satisfying $m_{\max}\leq R\leq M$ and substitutes it for $M$ in the row dimension of both shrink and expand launch grids, resulting in $L\lceil R/ B_M\rceil$ row tiles. The bound only changes the launch geometry. Blocks continue to use the existing adapter-grouped row-index array to gather input activations and scatter results to their original batch positions.

The launch additionally spans $S$ projection slices, $L$ adapter slots and $P$ split-$K$ partitions. Equation~\ref{eq:rowpulse-grid} gives the resulting three-dimensional launch grid:
\begin{equation}
    G_R=\left(
        P\left\lceil\frac{R}{B_M}\right\rceil
        \left\lceil\frac{N}{B_N}\right\rceil,\;
        S,\;L
    \right)
    \label{eq:rowpulse-grid}
\end{equation}
For shrink, $N$ is the LoRA rank; for expand, $N=D$ and $P=1$ since the expand kernel does not partition its reduction over the
LoRA rank across blocks.
We denote their stage-specific grids by $G_R^{\mathrm{shrink}}$ and $G_R^{\mathrm{expand}}$, respectively. On the shared-bound path, both use the same $R$ and $L$, with stage-specific tile settings. Row-bound specialization replaces $M$ with $R$ in the launch grid and its block-coordinate calculation, leaving the other grid dimensions and the underlying computation unchanged.

\subsubsection{Separating Row-Size Classes}
\label{sec:rowpulse-classes}

A shared row bound allows one large adapter group to enlarge every adapter's launch grid. This happens frequently when a batch contains both prefill and decode requests. To address this, we partition active adapters into two classes:
\begin{equation}
    \mathcal{A}_s=\{a:0<m_a\leq16\},\qquad
    \mathcal{A}_l=\{a:m_a>16\}
    \label{eq:rowpulse-classes}
\end{equation}
The small class uses row bound $R_s=16$, while the large class
uses $R_l=R$, where $R$ covers the largest active group.
These bounds specify row coverage, not tile height.
Shrink retains its shared-bound launch. We only apply class separation to expand as expand tiles span a much wider output dimension rather than the LoRA rank, so redundant row tiles are replicated across more output-column tiles, making expand a natural target for class separation.

To support fixed launch shapes for CUDA-graph replay, the small class retains the original $L$ adapter slots, marking nonmembers inactive. The large class uses a compact metadata list with a fixed capacity of $C=16$ slots. Each slot identifies an adapter.  Unused slots exit without matrix computation. 
For class $c$ with row bound $R_c$ and adapter-slot capacity $L_c$, the expand launch grid is
\begin{equation}
    G_c^{\mathrm{expand}}=
    \left(
        \left\lceil\frac{R_c}{B_M}\right\rceil
        \left\lceil\frac{D}{B_N}\right\rceil,\;
        S,\;L_c
    \right)
    \label{eq:rowpulse-class-grid}
\end{equation}
where $D$ is the projection's output width, $S$ is the number
of projection slices, and $B_M$ and $B_N$ are the row- and
column-tile sizes. With identical tile sizes, the shared-bound and separating class paths provision
\begin{equation}
    T_{\mathrm{shared}}=L\left\lceil\frac{R}{B_M}\right\rceil,
    \ 
    T_{\mathrm{class}}=
    L\left\lceil\frac{16}{B_M}\right\rceil+
    C\left\lceil\frac{R}{B_M}\right\rceil
    \label{eq:rowpulse-class-tiles}
\end{equation}
Class separation targets batches containing many small adapter groups and a few large ones, as can arise when decode requests share a batch with prefill requests. A shared bound lets the larger prefill groups inflate the expand grids of smaller decode groups. Separating them reduces this over-provisioning, but adds a kernel launch and metadata preparation. We therefore enable class-separated expand only when $|\mathcal{A}_s|\geq8$ and $1\leq|\mathcal{A}_l|\leq C$. The small-group threshold heuristically targets batches with sufficient opportunity to optimize the batch performance. Other batches, including capacity overflow, use the
shared-bound path without dropping any adapter work.

% Every active adapter belongs to exactly one class, whose
% row bound covers all its assigned rows. Both class launches
% preserve the original grouped-row mapping, so class separation
% neither omits nor duplicates LoRA corrections.
% Tile tuning is separate from this mechanism: our class-count
% ablation holds expand tile settings fixed and leaves shrink
% unchanged.

\subsubsection{Selecting and Replaying Row-Bucket Graphs}
\label{sec:rowpulse-graphs}

Adapting the row bound also requires accommodating CUDA graphs. vLLM captures GPU operations for a fixed execution shape and replays that sequence on subsequent decoding steps to reduce CPU launch overhead. Because a captured graph fixes its kernel launch dimensions, a single graph cannot directly accommodate arbitrary values of $R$.
We therefore capture a small set of row-bound and class-layout variants. The configured class policy is fixed during engine initialization. For padded batch size $M$, the supported bounds are
\begin{equation}
    \mathcal{R}(M)
    =
    \{16\cdot 2^k \mid k\in\mathbb{Z}_{\geq0},\;16\cdot 2^k<M\}
    \cup\{M\}
    \label{eq:rowpulse-buckets}
\end{equation}
For example, a batch with $M=256$ uses bounds of 16, 32, 64, 128, or 256. Each supported bound corresponds to a pre-captured graph. Let $d$ denote the existing graph descriptor, including the padded batch shape and adapter-count configuration. We extend it with $R$ and a layout flag $z$: $z=0$ selects shared-bound execution, and $z=1$ selects class-separated expand. The corresponding graph is $\Gamma(d,R,z)$.
At each decoding iteration, the host computes the assigned token-row count $m_a$ for each adapter by summing the scheduled token counts of requests using that adapter. 
For a nonempty active adapter set, we select the smallest supported bound covering the current $m_{\max}$:
\begin{equation}
    R=\min\{r\in\mathcal{R}(M):r\geq m_{\max}\}
    \label{eq:rowpulse-dispatch}
\end{equation}
The host selects $z=1$ when the class-separation conditions in Section~\ref{sec:rowpulse-classes} hold, and $z=0$ otherwise.
The engine then replays $\Gamma(d,R,z)$ with current inputs and adapter metadata. Within replay, shrink and the selected expand launches execute in dependency order using their captured grids. Selecting a different bucket or layout therefore changes the launch shape without recapturing the graph; if the bucket
and existing descriptor remain unchanged, the same graph is reused.
If a valid row-bound descriptor is unavailable, the kernel retains the full-batch bound $M$. 
%The host selects the graph using scheduled per-adapter row counts. GPU metadata preparation constructs the class lists without repacking activations or the grouped-row index array.
Algorithm~\ref{alg:rowpulse-launch} summarizes this process.

\begin{algorithm}[t]
    \caption{\textsc{GridTrim} graph selection and execution.}
    \label{alg:rowpulse-launch}
    \footnotesize
    \begin{algorithmic}[1]
        \Require Graph descriptor $d$ with padded row count $M$,
        active adapter set $\mathcal{A}$,
        row lists $\{\mathcal{I}_a\}_{a\in\mathcal{A}}$
        \Statex \textit{Runtime graph selection}
        \If{$\mathcal{A}=\varnothing$}
            \State Execute the backbone-only path
            \State \Return
        \EndIf
        \For{$a\in\mathcal{A}$}
            \State $m_a\gets|\mathcal{I}_a|$
        \EndFor
        \State $m_{\max}\gets\max_{a\in\mathcal{A}}m_a$
        \State $R\gets\min\{r\in\mathcal{R}(M):r\geq m_{\max}\}$
        \State $z\gets$ whether the class eligibility rule holds
        \State Prepare adapter metadata, including class lists if enabled
        \State Replay $\Gamma(d,R,z)$ with current inputs and metadata
        \Statex \textit{Within replay, for each LoRA operation:}
        \State Run shrink with captured grid $G_R^{\mathrm{shrink}}$
        to compute the intermediate activations
        \If{$z=1$}
            \For{each configured class $c$}
                \State Run expand with captured grid $G_c^{\mathrm{expand}}$
            \EndFor
        \Else
            \State Run expand with captured grid $G_R^{\mathrm{expand}}$
        \EndIf
    \end{algorithmic}
\end{algorithm}

\subsection{StepTrim: Decode-Time adapter release}
\label{sec:steppulse}

Applying a LoRA adapter at every decoding step is unnecessary for some requests, but deciding when to release the adapter of a request is difficult. 
We propose \textsc{StepTrim} to determine when to permanently release a request's adapter and continue decoding with the backbone alone.  \textsc{StepTrim} supports two release policies: a profile-based training-free policy that uses an adapter-task specific fixed release length $k$, and a learned policy that makes request-specific decisions. Both policies retain LoRA adapter during prefill and preserve the existing KV cache at release. The profiled-based policy requires no training. It calibrates a Pareto-frontier of the release position $k$ and task accuracy on the validation dataset, and picks the position $k$ that matches the serving requirements. The learned policy exploits differences in release sensitivity among requests. Our evaluation shows its additional benefits are task-backbone dependent.

\subsubsection{Profiled Release Policy}
The profiled-based policy selects a fixed release length $k$ for each task-adapter pair. We evaluate candidate release lengths and their corresponding accuracies on the validation set, retaining LoRA during prefill and the first $k$ decoding steps before switching to backbone-only decoding. The existing KV cache is preserved at release. $k=0$ corresponds to prefill-only execution. We select $k$ according to the desired validation quality-efficiency criterion and freeze it before deployment. 
%For the P50 comparison, we chose the candidate with the smallest validation accuracy loss among those achieving at least 50\% decode skipping. 
All requests assigned to the same adapter use the same selected release length. This policy, used by \name{}-Training-free, requires offline profiling but no training.

\subsubsection{Learned Release Policy}

\paragraph{One-way adapter release}

Consider request \(i\), assigned to adapter \(a_i\). Let \(z_{i,t}\in\{0,1\}\) denote whether its adapter is active at decode step \(t\).  \(z_{i,0}=1\) as we always apply the adapter during prefill. After a decode forward pass, it reads the final hidden vector \(h_{i,t}\in\mathbb{R}^{H}\). Then we apply a small trained predictor
associated with the request's adapter to compute
\begin{equation}
s_{i,t}=w_{a_i}^{\mathsf T}h_{i,t}+b_{a_i}
\end{equation}
The adapter assignment for the next decode step is
\begin{equation}
z_{i,t+1}=z_{i,t}\,\mathbf{1}[s_{i,t}\geq\tau_{a_i}]
\end{equation}
where \(\tau_{a_i}\) is a calibration threshold selected on validation data. Once \(z_{i,t}\) becomes zero, it remains zero until the request terminates. The request then follows the backbone-only path, while other requests in the same batch may continue using their adapters. 

\paragraph{Labeling with paired continuations}

The desired training label is not whether LoRA changes the next token, but whether releasing it early harms the final task outcome. For each training request, we apply the corresponding LoRA adapter normally and sample reachable decode positions. Those positions can be changed based on the dataset's generation length. For the dataset used in our evaluation, we sample at the following positions.
\begin{equation}
   \mathcal{T}=\{0,1,2,3,7,15,31,63\}
\end{equation}
We also tried to increase the number of sampling points at different positions along with more complex architectures, but the improvement is marginal. 

At each position \(t\), we fork the generation state into two free-running continuations:

\textbf{ON}: retain the adapter from \(t\) through EOS;

\textbf{OFF}: disable the adapter from \(t\) through EOS.

Both begin from an identical state. The OFF branch generates tokens until termination without using an adapter. Consequently, its final outcome captures both the immediate effect of the decision and its downstream effect through changed tokens and attention states. 

Let \(Q(\cdot)\in\{0,1\}\) be the task evaluator, such as answer accuracy or SQL
execution correctness. We label a candidate release point as risky if continuing with LoRA
produces a correct answer, but disabling LoRA from that point onward produces an incorrect answer:
\begin{equation}
 y_{i,t}=\mathbf{1}[Q(Y^{\mathrm{ON}}_{i,t})=1\ \land
                         Q(Y^{\mathrm{OFF}}_{i,t})=0]
\end{equation}
Thus, the predictor learns to rank the risk of releasing an adapter at a position. The training label does not treat a release as harmful when both branches fail.

\paragraph{Lightweight risk-predictor training}

We train one linear predictor per adapter using only its training split. Because longer generations can contribute more sampled decoding points, averaging over all points would overweight such requests. We instead give every request equal total weight:
\begin{equation}
\mathcal{L}=
\frac{1}{N}\sum_{i=1}^{N}\frac{1}{|\mathcal{T}_i|}
\sum_{t\in\mathcal{T}_i}
\mathcal{L}_{\mathrm{BCE}}
\bigl(w^{\mathsf T}h_{i,t}+b,\;y_{i,t}\bigr)
+\frac{\lambda}{2}\lVert w\rVert_2^2
\end{equation}
Here, $N$ is the number of training requests, $\mathcal{T}_i$ contains the sampled decoding points reached by request $i$, and $\mathcal{L}_{\mathrm{BCE}}$ denotes binary cross-entropy between the sigmoid-transformed logit and the risk label. The final term regularizes the head weights, with $\lambda=0.1$ across all adapters and training-set sizes.

The trained predictor produces a risk score $s_{i,t}=w_{a_i}^{\mathsf T}h_{i,t}+b_{a_i}$. We permanently release request $i$'s adapter when $s_{i,t}<\tau_{a_i}$, where $\tau_{a_i}$ is an adapter-specific threshold. Larger thresholds permit more aggressive release. We select $\tau_{a_i}$ using validation data for calibration. Candidate thresholds are drawn from quantiles of the predictor's training scores. For each candidate, we generate complete responses on the validation split and measure accuracy loss relative to full LoRA and the fraction of decode adapter calls skipped. We retain the Pareto-optimal candidates and select an operating point according to the desired quality-efficiency trade-off, freezing the threshold before deployment.

\section{Evaluation}

\begin{table*}[t]
    \centering
    \caption{
        Test-split Pass@1 accuracy (\%). Full LoRA applies LoRA throughout prefill and decoding. \name-Training-free uses the profiled threshold $k$ on each validation split that gives at least 50\% calibration-time decode skipping. For \textsc{\name}-P25/P50/P75, each adapter's threshold $\tau_a$ is selected independently on the validation split to achieve at least 25\%/50\%/75\% calibration-time decode skipping, respectively. \textsc{\name}-P50 is used in the throughput test unless specified.
    }
    \vspace{2 mm}
    \label{tab:pulselora-quality}

    \small
    \renewcommand{\arraystretch}{1.18}
    \setlength{\tabcolsep}{5pt}

    \resizebox{\textwidth}{!}{%
    \begin{tabular}{@{}lcccccccc@{}}
        \toprule
        & \multicolumn{4}{c}{\textbf{Llama-3.2-3B}}
        & \multicolumn{4}{c}{\textbf{Mistral-7B-v0.3}} \\
        \cmidrule(lr){2-5}
        \cmidrule(lr){6-9}
        \textbf{Method}
        & ScienceQA & Spider & MedQA-4 & GSM8K
        & ScienceQA & Spider & MedQA-4 & GSM8K \\
        \midrule
        Backbone-only
        & 38.00 & 44.00 & 45.48 & 6.14
        & 74.00 & 57.06 & 47.45 & 10.54 \\

        PreFT~\cite{lanpouthakoun2026preft}
        & 82.60 & 59.96 & 50.35 & 35.86
        & 87.40 & 64.60 & 59.23 & 49.20 \\

        Prefill-Only
        & 52.60 & 58.25 & 45.95 & 37.55
        & 85.93 & 65.22 & 60.02 & 48.02 \\

        \midrule

        \textsc{LoRazor}-Training-free
        & 83.20 & 63.31 & 52.37 & 41.55
        & 88.33 & 64.44 & 59.60 & 48.02 \\

        \textsc{LoRazor}-P25
        & 88.40 & 63.54 & 52.50 & 41.58
        & 92.07 & 65.76 & 59.99 & 53.65 \\

        \textsc{LoRazor}-P50
        & 87.27 & 62.54 & 52.42 & 41.24
        & 90.73 & 65.76 & 59.99 & 52.59 \\

        \textsc{LoRazor}-P75
        & 84.13 & 61.51 & 52.74 & 39.68
        & 87.00 & 63.19 & 59.32 & 50.72 \\
         \midrule
        vLLM (Full LoRA)~\cite{kwon2023efficient}
        & 90.00 & 65.34 & 52.55 & 44.02
        & 92.93 & 68.44 & 59.47 & 54.33 \\

        S-LoRA~\cite{slora}
        & 90.00 & 64.99 & 52.55 & 44.35
        & 92.80 & 68.57 & 59.70 & 55.12 \\

        \bottomrule
    \end{tabular}%
    }
\end{table*}

\subsection{Models and Datasets}

We evaluate Llama-3.2-3B and Mistral-7B-v0.3 on mathematical reasoning (GSM8K~\cite{cobbe2021gsm8k}), text-to-SQL generation
(Spider~\cite{yu-etal-2018-spider}), medical question answering
(MedQA-4~\cite{jin2021disease}), and scientific question answering (ScienceQA~\cite{lu2022learn}), training one LoRA adapter per backbone-task pair.
Table~\ref{tab:datasets} summarizes the datasets. GSM8K and Spider reserves training examples for validation and retain the official test split. MedQA-4 uses four-option questions; ScienceQA uses text-only examples from the official splits. 
We report pass@1 accuracy from greedy generation, scored by numerical answer correctness for GSM8K, SQL execution correctness for Spider, and answer-choice correctness for MedQA-4 and ScienceQA. Output limits are 256 tokens for GSM8K and Spider, 128 for MedQA-4, and 320 for ScienceQA.

\begin{table}[H]
\vspace{-5mm}
    \centering
    \small
    \caption{Processed dataset sizes.}
    \vspace{2mm}
    \label{tab:datasets}
    \begin{tabular}{lrrr}
        \toprule
        Dataset & Training & Validation & Test \\
        \midrule
        GSM8K     & 6726  & 373   & 1319 \\
        Spider    & 6282  & 359   & 1034 \\
        MedQA-4   & 10178 & 636 & 1273 \\
        ScienceQA & 5000  & 250   & 500   \\
        \bottomrule
    \end{tabular}
    \vspace{-2mm}
\end{table}

\subsection{Implementation and Serving Setup}

We implement \name{} on vLLM v0.30.0 (released on Sept 22, 2026) and conduct all experiments on NVIDIA H100 80GB GPUs. Throughput workloads contain 32768 requests, equally divided among four tasks, with test examples sampled repeatedly as needed. To vary adapter diversity from 4 to 128, we replicate the four trained task adapters under distinct logical identities. At 128 adapters, each task adapter has 32 identities with identical weights but separate GPU allocations, each executed as a distinct adapter. Thus, weights are not shared across identities. Main experiments assign requests uniformly across identities, with skewed traffic evaluated separately (Sec~\ref{sec:skew}). The learned StepTrim predictor is trained on 128 requests per task--backbone pair. For \name-P25/P50/P75, we select thresholds $\tau$ that maximize validation accuracy while achieving at least 25/50/75\% decode skipping, respectively, and freeze them before testing. Realized test-time skipping may differ. Additional serving details appear in Appendix~\ref{app:serving-details}.

\subsection{Baselines}
For task accuracy, we compare \name\ against vLLM, S-LoRA, Prefill-only, and PreFT~\cite{lanpouthakoun2026preft}. vLLM/S-LoRA applies LoRA during prefill and every decoding step. Prefill-only applies LoRA only during prefill.  PreFT is similar to Prefill-only, but it uses specialized training so the adapter is trained to only used on prefill but not decode. We also compare against our own \name-Training-free (without GridTrim), which disables LoRA after the first $k$ decoding steps. The selection of $k$ is based on the validation dataset of each backbone-task pair, details can be found in Appendix~\ref{app:fixed-prefix}.
For throughput, we compare against vLLM~0.30.0's default multi-LoRA serving, which incorporates Punica-derived kernels for heterogeneous adapter batching and CUDA-graph execution, S-LoRA, and \name-Training-free. We exclude prefill-only execution and PreFT from throughput comparisons because they incur substantial accuracy losses on some tasks and offer no flexibility for accuracy-throughput trade-off.

\subsection{Downstream Task Performance}

Table~\ref{tab:pulselora-quality} reports test-split Pass@1 accuracy across four tasks and two backbones. At the P50 operating point used in the main throughput evaluation, \name{} remains within 2.80 percentage points of full-LoRA accuracy on every backbone--task pair. It achieves higher accuracy than prefill-only execution on seven of eight pairs and PreFT on all eight. Compared with \name-Training-free, \name-P50 achieves higher accuracy on six of eight pairs, including all four Mistral tasks. Both policies are calibrated to achieve at least 50\% validation-time decode skipping, using an adapter-specific release length for the training-free policy and a predictor threshold for the learned policy. Their realized test-time skipping rates can differ. Appendices~\ref{app:fixed-prefix} and~\ref{app:test-skipping} provide the selected release lengths and P50 test-time skipping rates, respectively. Table~\ref{tab:release-policy-2pp} further compares the policies under a common validation accuracy loss budget.
Increasing the validation-time skipping target from P25 to P75 generally reduces accuracy, providing adjustable operating points for the quality-efficiency trade-off.

\subsection{Throughput Improvement}
Throughput experiments use the validation-selected P50 operating point unless otherwise specified. We evaluate both offline serving, where all requests are
available at the start, and open-loop serving, where requests arrive according to a Poisson process. We construct workloads from the test splits of the four tasks and assign requests to 4-128 logical adapters. Under uniform traffic, each adapter receives an equal share of requests over the workload, although individual runtime batches may be imbalanced.

\begin{figure}[t]
  \centering
  \includegraphics[width=\columnwidth]{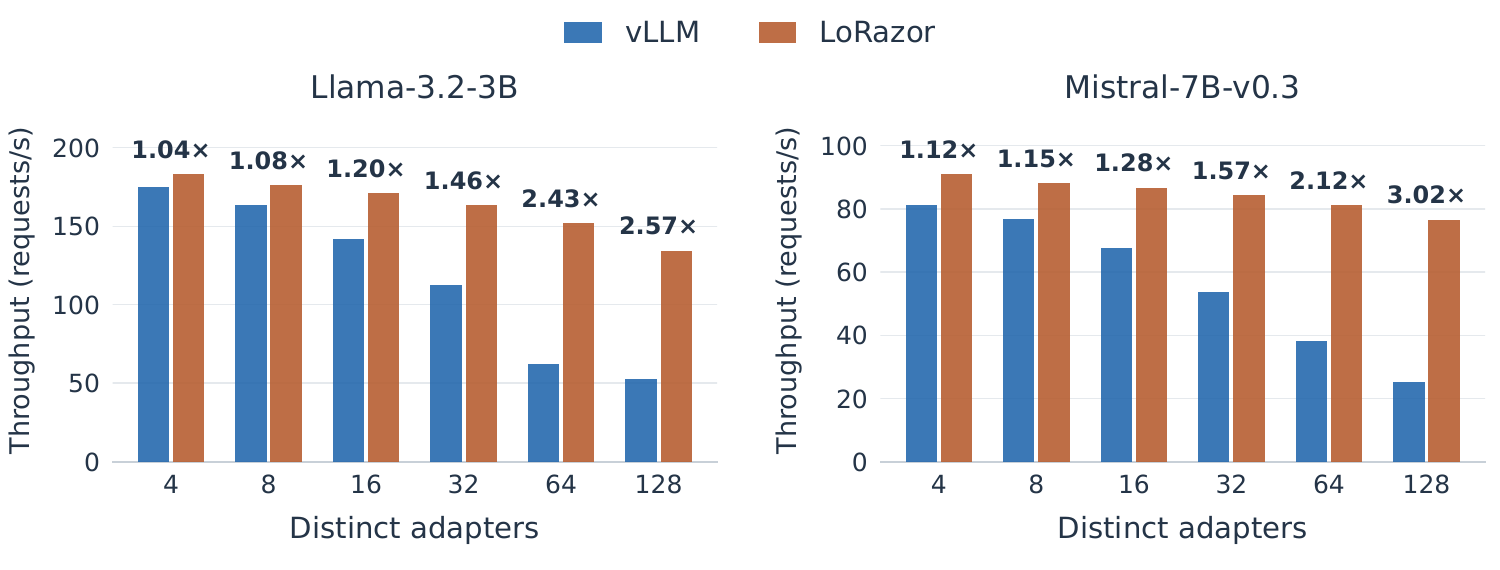}
  \vspace{-7 mm}
  \caption{Throughput versus adapter diversity under uniform traffic. Labels show \name's speedup over vLLM.}
  \vspace{-3 mm}
  \label{fig:throughput-scaling}
\end{figure}
\begin{figure}[t]
  \centering
  \includegraphics[width=\columnwidth]{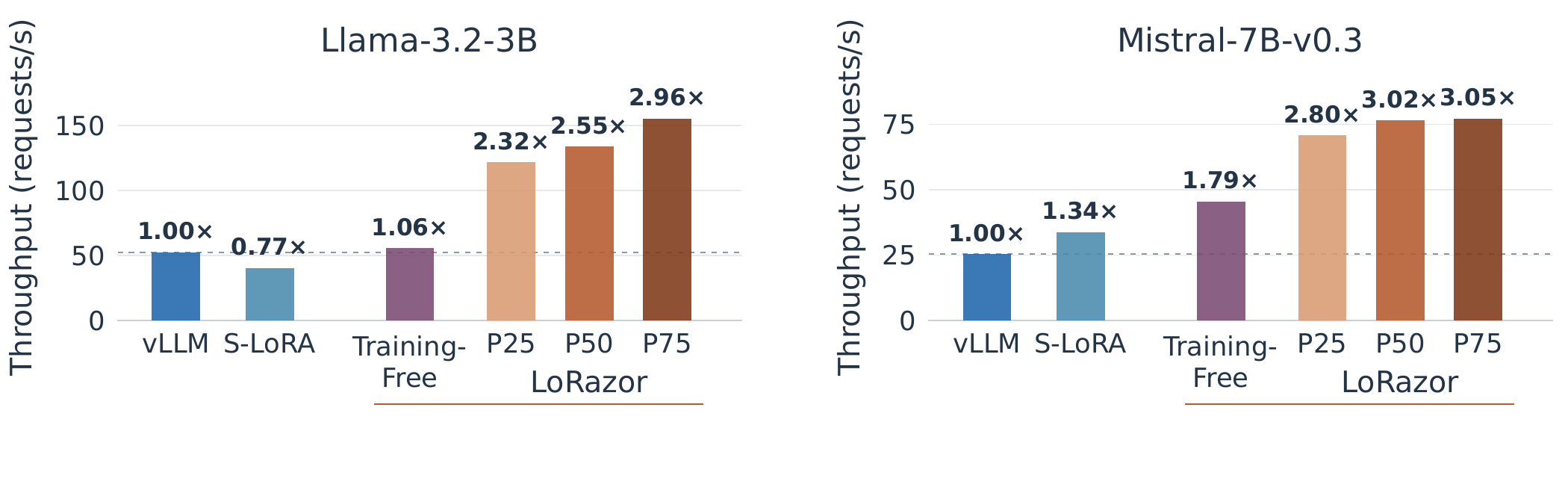}
  \vspace{-7 mm}
  \caption{Throughput with 128 adapters under uniform traffic.
  Labels show speedup over vLLM. Training-free uses default vLLM kernels; P25/P50/P75 use GridTrim.}
  \label{fig:throughput-baselines}
  \vspace{-3 mm}
\end{figure}

\begin{figure}[t]
  \centering
  \includegraphics[width=\columnwidth]{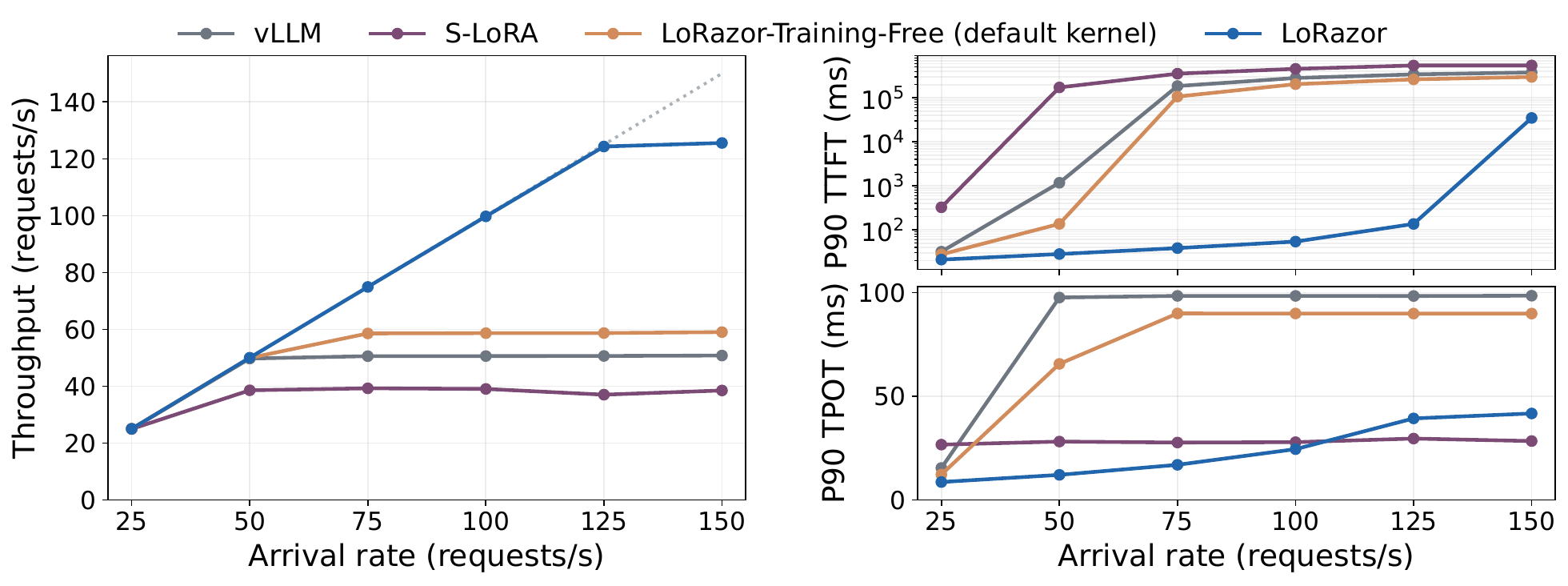}
  \vspace{-7 mm}
  \caption{Open-loop serving on Llama-3.2-3B under Poisson
arrivals with 128 adapters under uniform traffic: achieved throughput (left) and P90 TTFT and TPOT (right).}
  \label{fig:open-loop}
  \vspace{-3 mm}
\end{figure}
\paragraph{Offline throughput.}
Figure~\ref{fig:throughput-scaling} shows that vLLM's throughput decreases as adapter diversity increases, consistent with the rising per-step decoding latency in Figure~\ref{fig:serving-motivation}. \name{} reduces this latency by selectively disabling adapter execution and sizing the remaining LoRA kernel launches to the active adapter workload. As adapter diversity increases from 4 to 128, its throughput speedup over vLLM grows from 1.04$\times$ to 2.57$\times$ on Llama-3.2-3B and from 1.12$\times$ to 3.02$\times$ on Mistral-7B-v0.3. Figure~\ref{fig:throughput-baselines} compares the systems at 128 logical adapters. \name{} achieves 2.55$\times$ and 3.02$\times$ vLLM's request throughput on Llama and
Mistral, respectively, compared with 1.06$\times$ and 1.79$\times$ for Training-free (without GridTrim), and 0.77$\times$ and 1.34$\times$ for the S-LoRA compatibility port. Although \name-P50 produces slightly shorter responses on average, its output-token throughput also increases by 2.40$\times$ and 2.53$\times$ over vLLM, respectively (Table~\ref{tab:throughput-output-statistics}).
To isolate release-policy benefits, we additionally compare profiled-based (Training-free) and learned policy with GridTrim enabled for both under the same 2-percentage-point validation loss budget in Sec~\ref{sec:steppulse_deep}.

\paragraph{Open-loop serving.}
Figure~\ref{fig:open-loop} evaluates Llama-3.2-3B under Poisson arrivals ranging from 25 to 150 requests/s with uniform adapter assignment. We report throughput and the P90 time to first token
(TTFT) and time per output token (TPOT). \name{} closely tracks the arrival rate through 125 requests/s, whereas vLLM, the \name-Training-free policy (without GridTrim), and S-LoRA plateau at approximately 51, 59, and 39 requests/s, respectively. At 100 requests/s, \name{} maintains a P90 TTFT of approximately 55 ms and P90 TPOT of 25 ms, while vLLM experiences substantial queueing and a P90 TPOT near 99 ms. At 150 requests/s, \name{} throughput plateaus near 126 requests/s and its P90 TTFT rises sharply, indicating overload. These results show that the offline throughput gains translate into a higher supported arrival rate before substantial queueing occurs.

\begin{table}[t]
    \centering
    \small
    \setlength{\tabcolsep}{2pt}
    \caption{Output statistics for throughput experiments in Figure~\ref{fig:throughput-baselines}.}
    \vspace{2mm}
    \label{tab:throughput-output-statistics}
    \begin{tabular}{lcccc}
        \toprule
        & \multicolumn{2}{c}{Llama-3.2-3B}
        & \multicolumn{2}{c}{Mistral-7B-v0.3} \\
        \cmidrule(lr){2-3} \cmidrule(lr){4-5}
        Method
        & {\small $10^3$ tok/s}
        & {\small Tok/req}
        & {\small $10^3$ tok/s}
        & {\small Tok/req} \\
        \midrule
        vLLM
        & 2.79 & 53.33
        & 1.65 & 65.31 \\
        S-LoRA
        & 2.14 & 53.23
        & 2.21 & 65.33 \\
        LoRazor-Training-free
        & 2.72 & 48.99
        & 2.51 & 55.16 \\
        LoRazor-P25
        & 6.23 & 51.23
        & 4.20 & 59.26 \\
        LoRazor-P50
        & 6.71 & 50.19
        & 4.18 & 54.65 \\
        LoRazor-P75
        & 7.77 & 50.02
        & 4.14 & 53.55 \\
        \bottomrule
    \end{tabular}
    \vspace{-4mm}
\end{table}

\subsection{\textsc{StepTrim} and \textsc{GridTrim} Overhead}

\begin{figure}[h]
  \centering
  \includegraphics[width=\columnwidth]{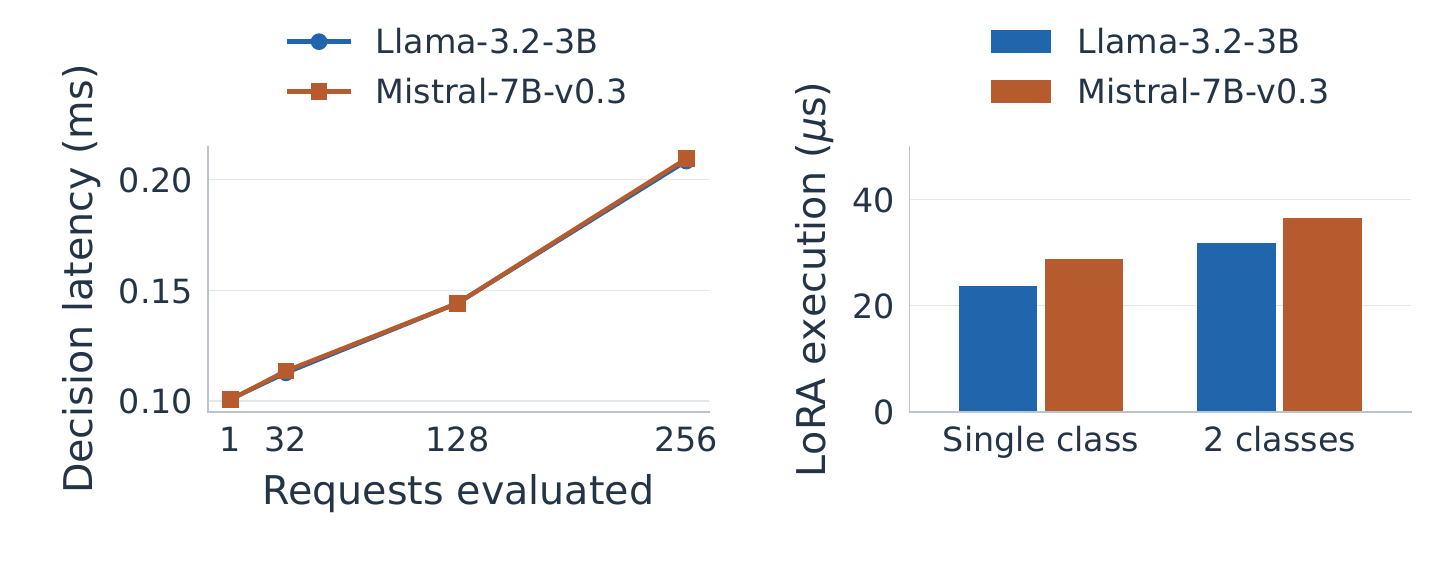}
  \vspace{-9 mm}
\caption{Left: StepTrim's 95th-percentile end-to-end release-decision latency with 128 predictor heads. Right: LoRA execution latency with single-class
and two-class GridTrim.}
  \vspace{-6 mm}
\label{fig:pulselora-overhead}
\end{figure}

Figure~\ref{fig:pulselora-overhead} reports StepTrim's release-decision latency and LoRA execution latency under single-class and two-class GridTrim. For StepTrim, we measure the learned policy's decision overhead. The profiled policy only checks whether each request has reached its preselected release step and requires no predictor evaluation. The StepTrim measurement includes hidden-state gathering, predictor evaluation, decision transfer, and release-state updates. We keep 128 adapter-specific predictor heads resident and vary the number of evaluated requests from 1 to 256; each request uses only its assigned adapter's head, which may be shared by multiple requests. The 95th-percentile decision latency increases from approximately
0.10 ms to 0.21 ms, about 1\% of a 20 ms decode step. Two-class execution adds metadata preparation and an additional expand launch, but the measured overhead is on the microsecond scale. Aside from latency overhead, GridTrim's additional row-bound and class-layout graph variants increase GPU memory consumption during capture by 1.61 GB relative to vLLM in our Llama-3.2-3B configuration. These graphs are captured during initialization and reused during decoding. Appendix~\ref{app:predictor_training} evaluates the offline cost of collecting paired continuations and fitting the predictor, along with the effect of training-set size on task accuracy.
\section{Ablation}
\subsection{Component ablation}
\begin{figure}[h]
  \centering
  \vspace{-2 mm}
  \includegraphics[width=\columnwidth]{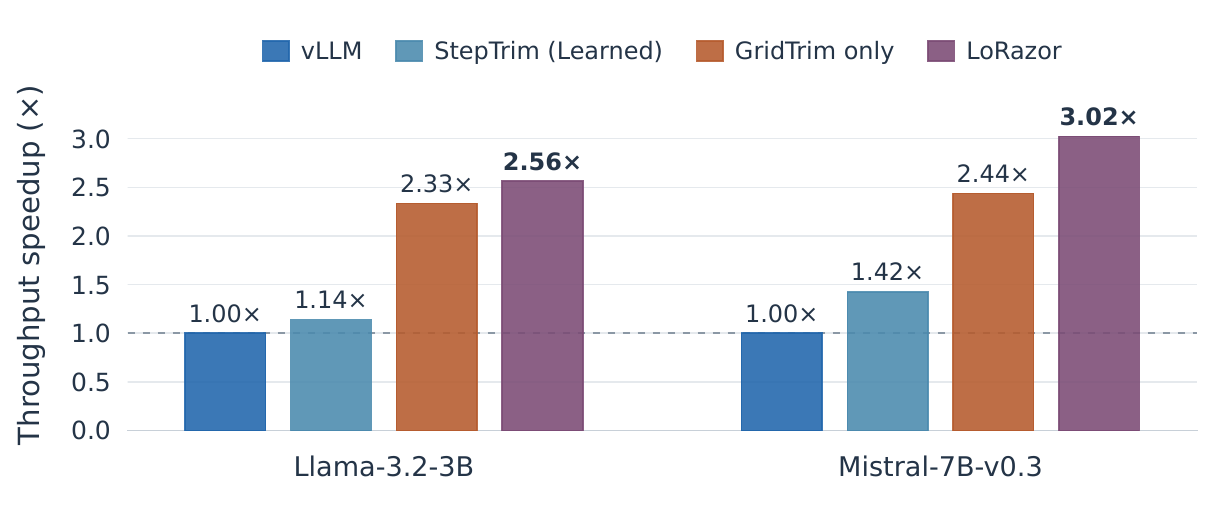}
  \vspace{-8 mm}
\caption{Component ablation under uniform traffic with 128 logical adapters. StepTrim and GridTrim both contribute to the improved performance.}
\vspace{-2 mm}
\label{fig:component-ablation}
\end{figure}
Figure~\ref{fig:component-ablation} isolates the contributions of
StepTrim and GridTrim under uniform traffic with 128 logical adapters. GridTrim alone achieves throughput improvements of $2.33\times$ on Llama-3.2-3B and $2.44\times$ on Mistral-7B-v0.3 over vLLM, while StepTrim with learned policy achieves $1.14\times$ and $1.42\times$, respectively.
Combining both components increases throughput to $2.56\times$ and $3.02\times$. The combined speedups should be interpreted together with the task accuracies in Table~\ref{tab:pulselora-quality} (\name-P50). StepTrim can provide additional savings even when GridTrim's row bound is already at its smallest bucket. Once all scheduled requests assigned to an adapter have released it, that adapter requires no further computation in the batch. If the active adapter count crosses an adapter-count specialization boundary, the engine can also select a graph with fewer adapter slots. Thus, release can reduce execution cost without further shrinking the row bound.

\subsubsection{GridTrim deepdive}
\begin{figure}[h]
  \centering

  \includegraphics[width=\columnwidth]{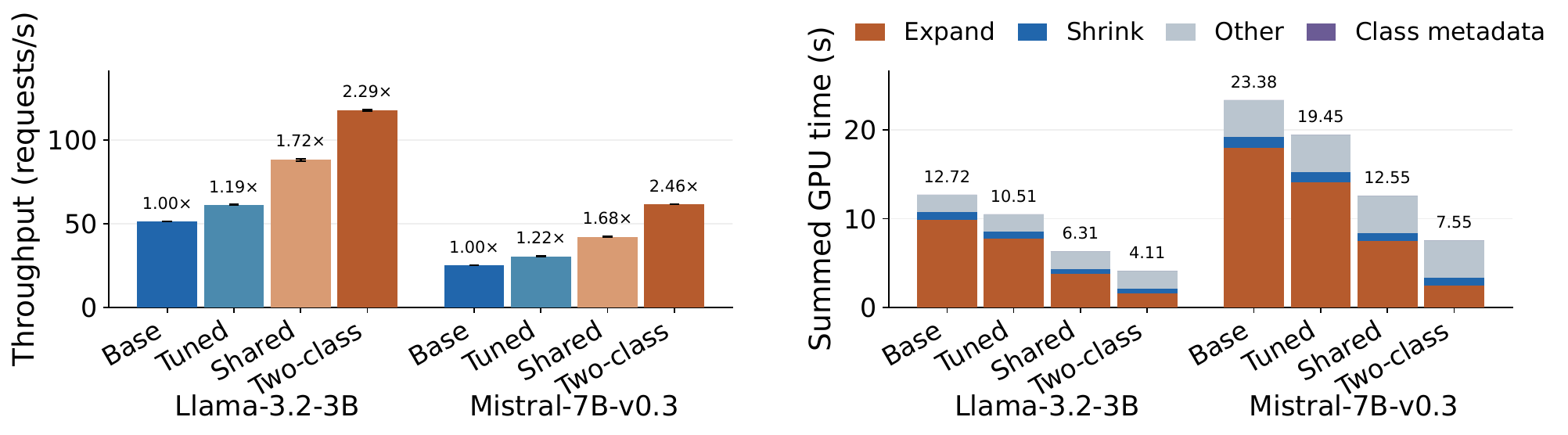}
  \vspace{-8 mm}
\caption{Left: throughput with tile tuning, shared row bounds, and two-class expand added progressively. Right: summed GPU execution time over the first 128 scheduling iterations.}
\label{fig:rowpulse-ablation}
\end{figure}

We further decompose GridTrim's gains in Figure~\ref{fig:rowpulse-ablation}, with StepTrim disabled. Tile tuning alone improves throughput by $1.19\times$ on Llama and $1.22\times$ on Mistral over the baseline. Holding these tile settings fixed, shared-row bounds increase throughput by another 43.7\% and 38.1\%, while two-class expand adds a further 33.5\% and 46.5\%, respectively. Overall, two-class GridTrim achieves $1.92\times$ and $2.02\times$ the throughput of tuned full-batch execution. The GPU-time breakdown over the first 128 scheduling iterations shows that class separation primarily reduces expand time, while shrink and other GPU work remain largely unchanged. This is expected as the two-class partition is applied during the expand stage.

\subsubsection{StepTrim deepdive: Learned or Profiled?}
\label{sec:steppulse_deep}
\begin{table}[h]
\vspace{-4 mm}
\centering
\small
\setlength{\tabcolsep}{2pt}
\caption{Test accuracy and decode LoRA skipping for policies selected under a 2 percentage point validation accuracy-loss budget.}
\vspace{2mm}
\label{tab:release-policy-2pp}
\begin{tabular}{llcc}
\toprule
& & \multicolumn{2}{c}{Accuracy / Skip (\%)} \\
\cmidrule(lr){3-4}
Model & Task & \name-Training-free & \name \\
\midrule
Llama
& ScienceQA & 87.60 / 25.46 & \textbf{88.80} / \textbf{56.96} \\
& Spider    & 62.86 / 73.94 & \textbf{62.96} / \textbf{76.27} \\
& MedQA-4   & \textbf{53.02} / \textbf{85.23} & 52.40 / 77.22 \\
& GSM8K     & 43.52 / 30.56 & 41.24 / 43.82 \\
\midrule
Mistral
& ScienceQA & 91.60 / 48.62 & 91.60 / 32.96 \\
& Spider    & 64.22 / 57.73 & \textbf{64.60} / \textbf{96.77} \\
& MedQA-4   & 59.78 / 94.15 & \textbf{60.17} / \textbf{98.53} \\
& GSM8K     & 53.98 / 43.59 & 53.90 / 47.64 \\
\bottomrule
\end{tabular}
\end{table}

We compare \name\ with \name-Training-free on each backbone-task pair. Both use \textsc{GridTrim} and are selected under a 2-percentage-point validation accuracy loss budget relative to full-LoRA execution. The selected policies are frozen before test evaluation. The budget applies to validation accuracy, realized test accuracy may not match.
Table~\ref{tab:release-policy-2pp} compares the two policies across all eight backbone--task pairs. \name\ achieves higher observed accuracy and decode skipping on four pairs. The largest skipping gains occur on Llama-ScienceQA and Mistral-Spider, where skipping increases by 31.50 and 39.04 percentage points, respectively, alongside accuracy improvements of 1.20 and 0.38 points. Gains are smaller on Llama-Spider and Mistral-MedQA-4. However, \name-Training-free achieves higher accuracy and skipping on Llama-MedQA-4 and higher skipping at equal accuracy on Mistral-ScienceQA. Both GSM8K pairs exhibit accuracy-skipping trade-offs. These results show that the benefits of learned policy are task-dependent and that profile-based release remains competitive. Appendix~\ref{app:release-cutoffs} examines request-level release-cutoff distributions to help explain these differences.

\subsection{Traffic distribution and row-size classes}
\label{sec:skew}

\begin{figure}[h]
  \centering
  \includegraphics[width=\columnwidth]{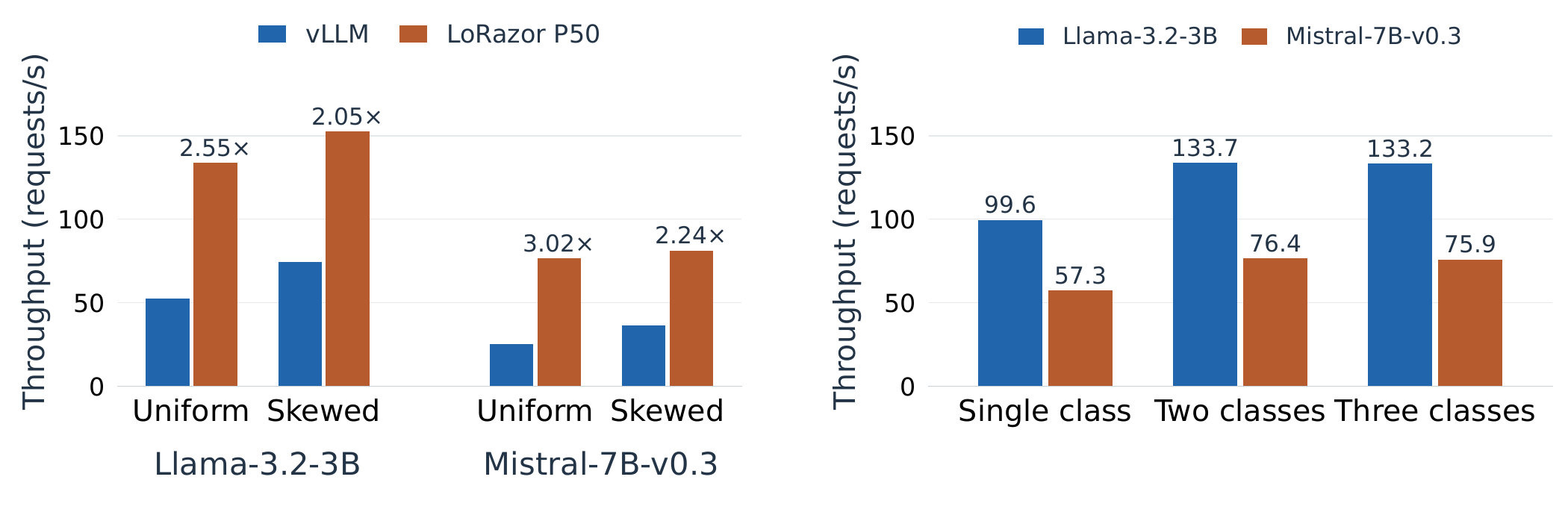}
  \vspace{-6mm}
  \caption{Effects of traffic distribution (left) and row-size classes (right) on throughput with 128 logical adapters.}
  \label{fig:traffic-ablation}
\end{figure}

Figure~\ref{fig:traffic-ablation} examines the effects of adapter popularity and class-separated execution with 128 logical adapters. Uniform traffic distributes requests evenly across adapters over the workload, whereas skewed traffic concentrates requests on a subset of adapters. Under skewed traffic, 24 of the 128 logical adapters receive approximately 80\% of requests. Appendix~\ref{app:serving-details} details the workload construction. \name\ achieves speedups over vLLM of $2.55\times$ and $3.02\times$ under uniform traffic on Llama-3.2-3B and Mistral-7B-v0.3, respectively, compared with $2.05\times$ and $2.24\times$ under skewed traffic. Both systems achieve higher absolute throughput under skewed traffic. The smaller relative speedup reflects vLLM's improvement as well. Concentrating requests on popular adapters can improve per-adapter batching and reduce the number of active adapters within individual batches. For \name, class-separated execution also limits the impact of
large adapter groups on smaller groups' launch grids.
The right panel isolates the effect of row-size classes by holding tile settings fixed and leaving shrink unchanged. Separating expand into two classes increases throughput from 99.6 to 133.7 requests/s on Llama and from 57.3 to 76.4 requests/s on Mistral, improvements of approximately 34\% and 33\%. Adding a third class ($  64 \ge m_{a} > 16$) yields 133.2 and 75.9 requests/s, respectively, providing no further improvement in this experiment. We therefore use two classes in the main evaluation.

\section{Related Work}
\label{sec:related-work}

\subsection{Selective Adapter Execution}

Prior work reduces adaptation cost by restricting where adapters operate or selecting which adapters contribute. AdapterDrop~\cite{ruckle2021adapterdrop} reduces execution across model depth by removing adapters from lower Transformer layers, while ReFT~\cite{wu2024reft} learns interventions on selected hidden representations. PrAd~\cite{ma2025prad} and PreFT~\cite{lanpouthakoun2026preft} restrict adaptation to prompt tokens through specialized training, eliminating adapter computation during decoding. A complementary direction dynamically routes or combines adapters at the token level. Arrow~\cite{arrow}, X-LoRA~\cite{xlora}, and LoRA Mixer~\cite{lora-mixer} select or weight adapter contributions with routing applied across adapters, layers, or projection matrices. AdaFuse~\cite{li2026adafuse} further combines token-level pre-gating with fused execution to reduce routing and kernel-launch overhead. Beyond adapters, SkipDecode~\cite{del2023skipdecode} and LayerSkip~\cite{elhoushi2024layerskip} reduce backbone computation through layer skipping or early-exit mechanisms. \name\ instead uses release policies to determine when requests can permanently switch from adapters to backbone-only decoding with limited impact on task accuracy.

\subsection{Multi-LoRA Serving and Kernel Optimization}

Multi-LoRA serving systems improve efficiency through heterogeneous batching, memory management, and scheduling. PetS~\cite{zhou2022pets} supports heterogeneous parameter-efficient tasks through shared backbone operators and coordinated batching. FLoRA~\cite{wen2024batched} allows each input example to use distinct low-rank adaptation weights. Punica~\cite{chen2024punica} enables heterogeneous adapter batching through specialized kernels and consolidates workloads across GPUs. S-LoRA~\cite{slora} combines heterogeneous batching with unified paging of adapter weights and KV caches. dLoRA~\cite{dlora} dynamically chooses merged and unmerged execution and coordinates request-adapter migration. Toppings~\cite{li2025toppings} overlaps adapter loading with CPU-assisted prefill. Chameleon~\cite{iliakopoulou2025chameleon} combines adapter caching with adapter-aware scheduling. LoRAServe~\cite{jaiswal2025serving} optimizes distributed adapter placement and routing. These systems improve where and how requested adapters execute. \name\ on the other hand decides how long each request uses its adapter during the decode generation, and releases the adapter accordingly to reduce the adapter usage in each batch iteration.

\subsection{Compact Adapter Representations}

Prior work reduces the memory footprint required for task specialization. VeRA~\cite{kopiczko2024vera} learns scaling vectors over shared frozen random projections. Compress then Serve~\cite{cts} jointly compresses trained adapters into shared bases and adapter-specific scaling matrices. SALT~\cite{li2026pin} trains shared domain centroids and ultra-low-rank task residuals to reduce per-adapter storage and transfer costs. Beyond LoRA, DeltaZip~\cite{yao2025deltazip} compresses weight differences between a shared backbone and fully fine-tuned models to support efficient concurrent serving. These methods change how adaptation is represented or stored to improve serving efficiency. \name\ complements this direction by reducing adapters usage in decode generation and adapting the computation kernel to the remaining workload.

\section{Discussion}

The choice between learned and profiled release policy of \name\ depends on the workload and backbone model. Although requests differ in their sensitivity to adapter release, a learned policy must identify favorable release points to exploit this variation. Currently, \name\ improves both accuracy and skipping on some backbone-task pairs, while others favor profiled release or exhibit accuracy-skipping trade-offs. 
Future work could investigate signals that better distinguish requests permitting early release from those benefiting from longer adapter execution. Such signals could improve prediction or support hybrid policies that combine profiled timing with request-specific decisions. Furthermore, both release policies rely on validation-based calibration and provide no per-request correctness guarantee. Workload shifts may require recalibration. Selective adapter reactivation is another direction, but must account for the KV state accumulated during backbone-only decoding and the additional execution cost.

\section{Conclusion}

We present \name, a multi-LoRA serving framework that enhances efficiency by strategically controlling the adapter execution duration during the decode stage. \textsc{StepTrim} employs release policies to permanently release adapters during decoding, boosting efficiency with marginal accuracy trade-offs. Meanwhile, \textsc{GridTrim} resizes kernel launches to match the remaining active adapter workload, eliminating unnecessary GPU block launches. Together, these techniques substantially improve multi-LoRA serving throughput with negligible impact on downstream task accuracy.
\clearpage

\bibliography{example_paper}
\bibliographystyle{mlsys2025}
\balance

\newpage
%\input{ai_usage}
% We used ChatGPT to polish the writing of the paper, gather the literature that is related to our work. Codex is used to help debugging the framework as well as set up the baseline repository for comparison purposes. All system diagrams shown in paper are hand drawn, no AI is used for image generation.

\clearpage
\appendix
\section{Appendix}

\subsection{Adapter execution during prefill}
\label{app:prefill}

\begin{figure}[h]
  \centering
  \includegraphics[width=\columnwidth]{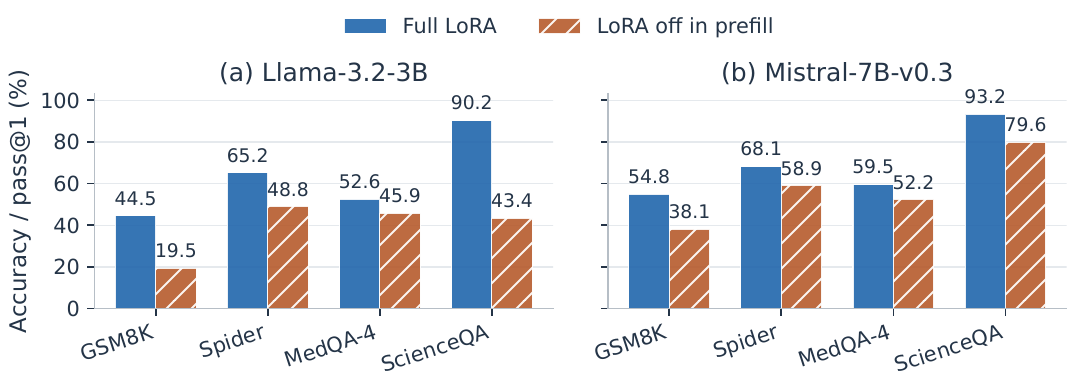}
  \vspace{-4mm}
  \caption{Disabling LoRA prefill significantly reduces task accuracy even when LoRA remains enabled throughout decoding.}
  \label{fig:prefill-ablation}
  \vspace{-2mm}
\end{figure}

The limited accuracy loss observed for selective adapter release during decoding does not extend to skipping LoRA for the entire prefill. Figure~\ref{fig:prefill-ablation} compares full-LoRA
execution (LoRA enabled throughout prefill and decoding) with a configuration that disables LoRA only during prefill while keeping it enabled throughout decoding.

Removing LoRA from the entire prefill causes substantially larger downstream accuracy losses than many decode-release configurations. This experiment evaluates skipping the entire prefill; selectively skipping adapter execution within prefill remains unexplored.

Based on these results, \name\ keeps adapters enabled throughout
prefill and focuses on request-specific release during decoding.
StepTrim decides when each request releases its adapter, while
GridTrim efficiently executes the remaining active adapter rows.

\subsection{Serving setup details}
\label{app:serving-details}

\paragraph{LoRA rank and target projections.}
For each backbone-task pair, we select the adapter with the highest validation Pass@1 accuracy among nine configurations spanning three target-module sets (Q/V, attention-only, and attention plus feed-forward projections) and three learning rates ($5\times10^{-5}$, $10^{-4}$, and $2\times10^{-4}$).
All adapters use rank $r=8$. For Llama-3.2-3B, GSM8K, Spider, and ScienceQA adapt the attention
projections $\{q_{\mathrm{proj}},k_{\mathrm{proj}}, v_{\mathrm{proj}},o_{\mathrm{proj}}\}$ and the feed-forward projections $\{\mathrm{gate}_{\mathrm{proj}}, \mathrm{up}_{\mathrm{proj}},\mathrm{down}_{\mathrm{proj}}\}$;
MedQA-4 adapts only the four attention projections. For Mistral-7B-v0.3, GSM8K, Spider, and ScienceQA adapt the four attention projections, whereas MedQA-4 adapts only $q_{\mathrm{proj}}$ and $v_{\mathrm{proj}}$.
Logical adapters replicate these task-specific weights under distinct adapter identities.

\paragraph{Skewed adapter traffic generation.}
The workload contains 32768 requests, equally divided among the four tasks. At 128 logical adapters, each task has 32 adapter identities. We randomly designate six identities per task as hot, obtained by rounding $0.2\times32$, and assign them 6554 of that task's 8192 requests, approximately 80\%.
The remaining 1638 requests are assigned to the other 26 identities. Thus, 24 of 128 adapters receive approximately 80\% of the workload. Only adapter assignments change between uniform and skewed traffic; prompts and request orders remain unchanged. 

\paragraph{Serving configuration and measurement.}
For vLLM and \name, we allow at most 256 concurrently scheduled sequences and 2048 scheduled tokens per iteration, with chunked prefill enabled and prefix caching disabled. We reserve 32 GB for the KV cache and configure capacity for 128 resident adapters. We enable full CUDA graphs and adapter-count specialization, and disable asynchronous scheduling.
Chunked prefill allows prompt chunks to share scheduling iterations with decoding requests, producing the heterogeneous adapter-group sizes targeted by two-class GridTrim.

For offline serving, we submit the complete request trace in a single generation call. Throughput is the number of completed requests divided by elapsed wall-clock time, including request admission and final draining but excluding initialization and warmup. Each measured run lasts at least 180 seconds. For open-loop serving, requests arrive according to a Poisson process at rates of 25-150 requests/s, independently of earlier completions. We report P90 TTFT and P90 TPOT in the open-loop serving along with the throughput.

\subsection{Profile-based release configuration}
\label{app:fixed-prefix}

Profile-based release applies LoRA during prefill and the first $k$ decoding steps, then disables it for the remaining generation. For each backbone-task pair, we select the candidate $k$ with the smallest validation accuracy loss among those achieving at least 50\% decode LoRA skipping. The selected values are frozen before test evaluation. This criterion matches the P50 skipping target, but may not realize the exact skipping rates on the test split. The value of $k$ used in the evaluation can be found in Table~\ref{tab:fixed-prefix-k}.

\begin{table}[h]
    \centering
    \caption{Validation-selected release lengths $k$. A value of zero denotes prefill-only execution.}
    \vspace{2mm}
    \label{tab:fixed-prefix-k}
    \small
    \renewcommand{\arraystretch}{1.1}
    \setlength{\tabcolsep}{4pt}
    \begin{tabular*}{\columnwidth}{@{\extracolsep{\fill}}lcc@{}}
        \toprule
        Task & Llama-3.2-3B & Mistral-7B-v0.3 \\
        \midrule
        GSM8K     & 32 & 0  \\
        Spider    & 8  & 16 \\
        MedQA-4   & 4  & 4  \\
        ScienceQA & 16 & 16 \\
        \bottomrule
    \end{tabular*}
\end{table}

\subsection{Realized test-time skipping rates}
\label{app:test-skipping}

P50 denotes a validation-time skipping target rather than a fixed
test-time skipping rate. We select a threshold on validation data and freeze it before
test evaluation. Table~\ref{tab:test-skipping} reports the resulting
fraction of decode LoRA skipped.

\begin{table}[h]
    \centering
    \caption{Test-time decode LoRA skipping rates (\%) for \name-P50.}
    \label{tab:test-skipping}
    \small
    \vspace{2mm}
    \renewcommand{\arraystretch}{1.1}
    \setlength{\tabcolsep}{4pt}
    \begin{tabular*}{\columnwidth}{@{\extracolsep{\fill}}lcc@{}}
        \toprule
        Task & Llama-3.2-3B & Mistral-7B-v0.3 \\
        \midrule
        GSM8K     & 64.15 & 58.39 \\
        Spider    & 69.48 & 60.68 \\
        MedQA-4   & 54.53 & 82.73 \\
        ScienceQA & 50.65 & 55.95 \\
        \bottomrule
    \end{tabular*}
\end{table}

Realized skipping varies across tasks because the selection criterion requires at least 50\% skipping on validation data, rather than exactly 50\%, and the selected policy can behave differently on the test split. These rates are weighted by decode length: longer responses contribute more calls to the denominator.
They therefore do not directly describe the fraction of requests released or their release positions. Our saved evaluation logs contain aggregate skipping counters but not per-request release steps, so release-position distributions are not reported.

\subsection{Predictor training cost}
\label{app:predictor_training}
\begin{figure}[h]
  \centering
  \includegraphics[width=\columnwidth]{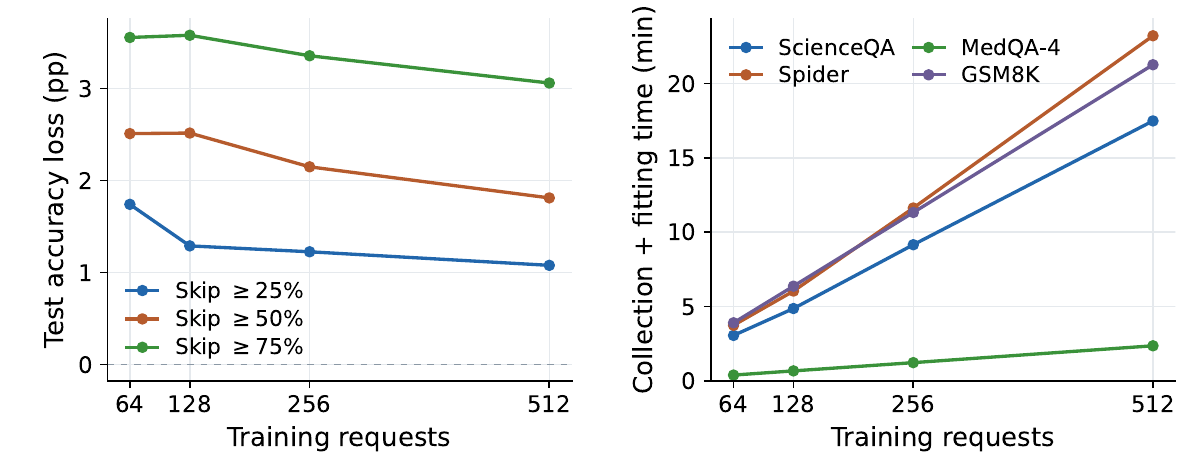}
  \vspace{-5mm}
\caption{Predictor quality changes modestly with additional training requests, while the one-time offline collection and fitting remain inexpensive.}
\label{fig:training-size}
\end{figure}

Figure~\ref{fig:training-size} shows how predictor quality and training cost scale with the number of training requests. In the left panel, each curve corresponds to a validation skipping target of at least 25\%, 50\%, or 75\% of decode LoRA calls, with a threshold selected separately at each training-set size. The vertical axis reports downstream test-accuracy loss. Increasing the training set from 128 to 512 requests yields only modest accuracy improvements (less than one percentage point). The right panel shows that data and label collection plus predictor fitting time scales approximately linearly with training-set size. This is a one-time offline cost incurred when preparing a risk predictor for a new adapter. The wall-clock cost varies across tasks because label collection generates paired continuations: longer outputs require more decoding, whereas MedQA-4 is substantially faster because its responses are shorter.

\subsection{Releasing threshold selection}

\begin{figure}[h]
  \centering
  \includegraphics[width=\columnwidth]{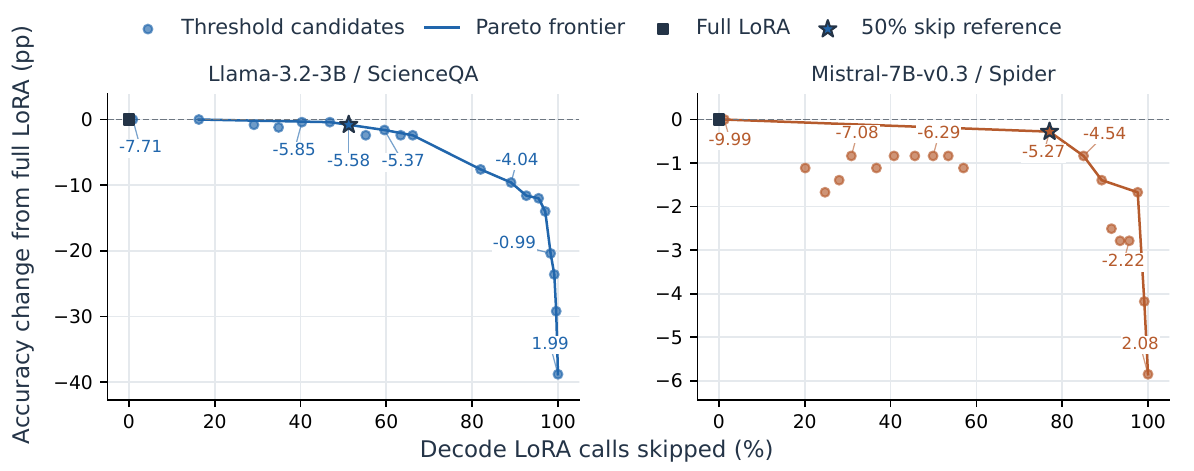}
  \vspace{-6 mm}
\caption{Release-threshold trade-offs Pareto frontier on validation data. Annotations show thresholds $\tau$. Stars mark the selected operating points achieving at least 50\% skipping.}
\label{fig:threshold-ablation}
\end{figure}

We vary the release threshold $\tau$ while holding the predictor checkpoint and validation workload fixed. Candidate thresholds are defined by quantiles of the predictor's scores on training data, covering conservative to aggressive release settings; each candidate is then evaluated on the same validation workload. Figure~\ref{fig:threshold-ablation} shows that more permissive thresholds increase decode skipping but can reduce accuracy. For Llama-3.2-3B on ScienceQA, a threshold of $-5.58$ skips 51.2\% of decode LoRA calls with only 0.8 percent point accuracy loss; more aggressive thresholds incur substantially larger losses. For Mistral-7B-v0.3 on Spider, a threshold of $-5.27$ enables 77.1\% skipping with only 0.3 percent point loss.
This demonstrates performance sensitivity to the release threshold and supports selecting an operating point separately for each adapter. For the reported throughput experiments on P50, we select the threshold with the smallest validation accuracy loss among Pareto frontier candidates achieving at least 50\% decode LoRA skipping on the validation data. We freeze this threshold before evaluating throughput on workloads constructed from the test data. 

\subsection{Profile-based or learned released policy?}
\label{app:release-cutoffs}

To understand the task-dependent benefits of learned release policy , we examine how release sensitivity varies across requests. This analysis complements the calibrated policy comparison in Table~\ref{tab:release-policy-2pp} by characterizing hindsight opportunities for request-specific release.

For each validation request, we first generate an answer with full-LoRA execution. We then repeat generation at every integer release cutoff $k$ within the tested horizon, retaining LoRA during prefill and the first $k$ decoding calls before permanently switching to backbone-only decoding. Thus, $k=0$ corresponds to prefill-only LoRA. Each continuation retains the KV cache produced before release and is scored using the task's evaluator. We restrict this analysis to requests answered correctly by full LoRA. For each request, we define its \emph{stable-safe cutoff} as the smallest tested cutoff at which the answer is correct and remains correct at every later tested cutoff. This definition accounts for nonmonotonic release sensitivity: a request may be answered correctly at an early cutoff but incorrectly at a later one. Consequently, the stable-safe cutoff need not be the earliest cutoff that produces a correct answer.

\begin{figure}[t]
    \centering
    \includegraphics[width=\columnwidth]{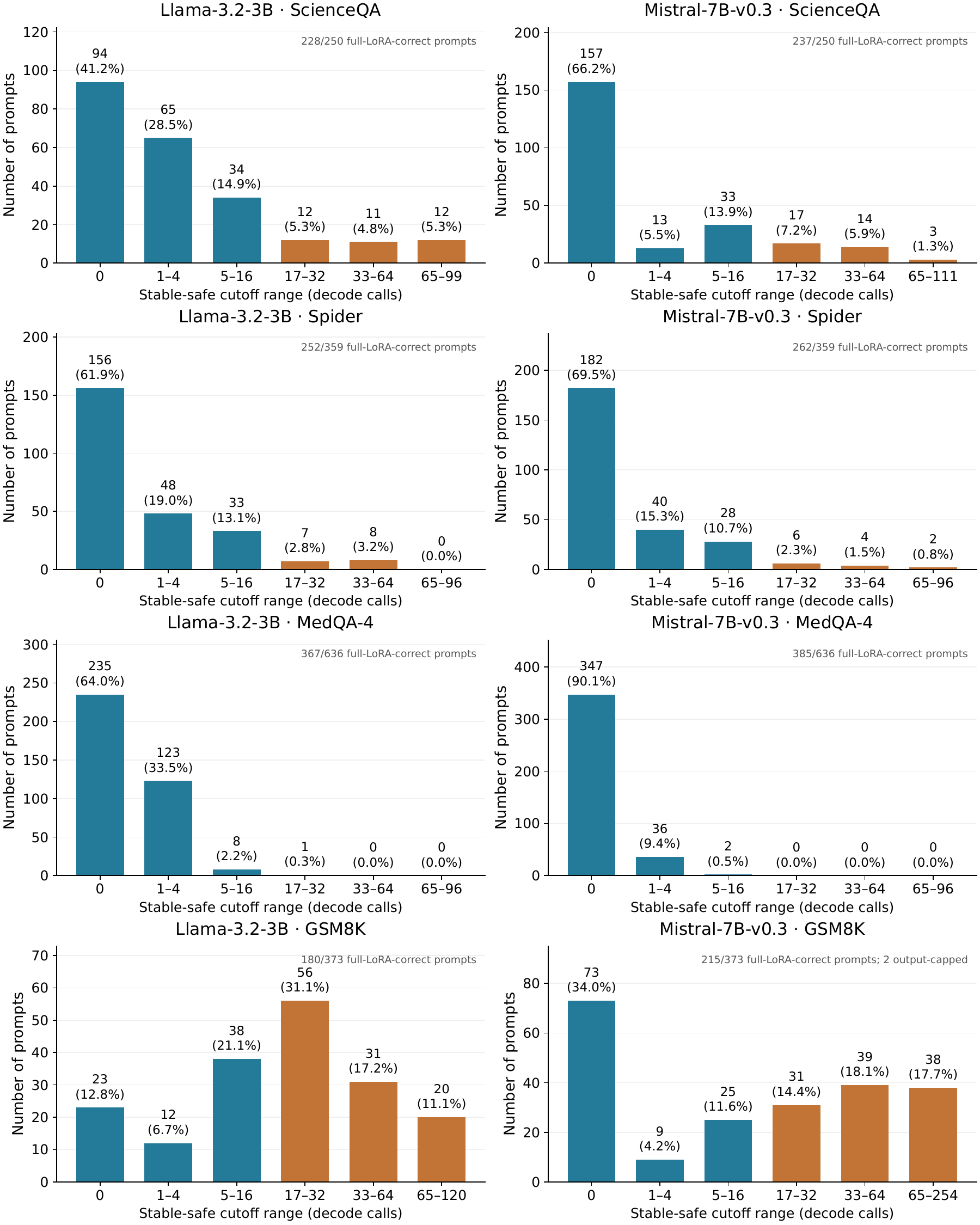}
    \caption{
    Distributions of stable-safe release cutoffs on validation requests answered correctly by full LoRA. A cutoff of $k$ retains LoRA during prefill and the first $k$ decoding calls.
    }
    \label{fig:release-cutoff-distributions}
\end{figure}

Figure~\ref{fig:release-cutoff-distributions} groups requests by their stable-safe cutoffs. Percentages are calculated among full-LoRA-correct requests within each backbone-task pair.

On MedQA-4, stable-safe cutoffs are concentrated in the first few decoding calls: 97.5\% of full-LoRA-correct Llama requests and 99.5\% of Mistral requests have cutoffs at most four.
This concentration helps explain why a profiled-based policy can be competitive, since there is relatively little variation in stable-safe cutoffs beyond the initial decoding steps.
In Table~\ref{tab:release-policy-2pp}, profile-based policy achieves both higher accuracy and higher skipping on Llama-MedQA-4. On Mistral-MedQA-4, profile-based policy already skips 94.15\% of decode LoRA execution, leaving limited additional skipping headroom. Learned policy increases skipping by 4.38 percentage points and accuracy by 0.39 points.

ScienceQA and Spider contain substantial populations with immediate-release opportunities alongside requests with later stable-safe cutoffs. On Llama-ScienceQA, 41.2\% of requests have cutoff zero, while 15.4\% have cutoffs greater than 16. A shared profile-based policy must balance retaining adapters for later-cutoff requests against unnecessary execution for requests that permit earlier release. The simultaneous improvements of the learned policy in accuracy and skipping on Llama-ScienceQA are consistent with exploiting this variation.

Spider also contains large immediate-release populations: 61.9\% of Llama requests and 69.5\% of Mistral requests have cutoff zero. Although most remaining requests reach their stable-safe cutoff within 16 calls, a minority have later cutoffs. Learned policy improves both accuracy and skipping on both Spider pairs, with a substantially larger skipping improvement on Mistral.

GSM8K exhibits substantially later stable-safe cutoffs: 59.4\% of full-LoRA Llama requests and 50.2\% of Mistral requests have cutoffs greater than 16. The learned policy yields higher skipping but lower observed accuracy, with a 2.28-percentage-point accuracy decrease on Llama and a 0.08-point decrease on Mistral. These results illustrate accuracy-skipping trade-offs rather than simultaneous gains.

%%%%%%%%%%%%%%%%%%%%%%%%%%%%%%%%%%%%%%%%%%%%%%%%%%%%%%%%%%%%%%%%%%%%%%%%%%%%%%%
%%%%%%%%%%%%%%%%%%%%%%%%%%%%%%%%%%%%%%%%%%%%%%%%%%%%%%%%%%%%%%%%%%%%%%%%%%%%%%%
% SUPPLEMENTAL CONTENT AS APPENDIX AFTER REFERENCES
%%%%%%%%%%%%%%%%%%%%%%%%%%%%%%%%%%%%%%%%%%%%%%%%%%%%%%%%%%%%%%%%%%%%%%%%%%%%%%%
%%%%%%%%%%%%%%%%%%%%%%%%%%%%%%%%%%%%%%%%%%%%%%%%%%%%%%%%%%%%%%%%%%%%%%%%%%%%%%%
% \appendix
% \section{Please add supplemental material as appendix here}
% %
% Put anything that you might normally include after the references as an appendix here, {\it not in a separate supplementary file}. Upload your final camera-ready as a single pdf, including all appendices.

%%%%%%%%%%%%%%%%%%%%%%%%%%%%%%%%%%%%%%%%%%%%%%%%%%%%%%%%%%%%%%%%%%%%%%%%%%%%%%%
%%%%%%%%%%%%%%%%%%%%%%%%%%%%%%%%%%%%%%%%%%%%%%%%%%%%%%%%%%%%%%%%%%%%%%%%%%%%%%%

\end{document}